\documentclass[runningheads]{llncs}
\usepackage[T1]{fontenc}
\usepackage{pifont}
\newcommand{\cmark}{\ding{51}}%
\newcommand{\xmark}{\ding{55}}%
\usepackage{graphicx}
\usepackage{amsfonts}
\usepackage{amssymb}
\usepackage{subfig}
\usepackage{multirow}
\usepackage{mathtools}
\usepackage{hyperref}
\begin{document}
\title{Segmentation by registration-enabled SAM prompt engineering using five reference images}

%
\titlerunning{Registration-enabled SAM prompt engineering}
%

\author{Yaxi Chen\inst{1}\orcidID{0009-0007-5906-899X} \and
Aleksandra Ivanova\inst{2}\orcidID{0009-0000-4113-8928} \and
Shaheer U. Saeed\inst{3}\orcidID{0000-0002-5004-0663} \and
Rikin Hargunani\inst{4}\orcidID{0000-0002-0953-8443} \and
Jie Huang\inst{1}\orcidID{0000-0001-7951-2217} \and
Chaozong Liu\inst{2,4}\orcidID{0000-0002-9854-4043} \and
Yipeng Hu\inst{3}\orcidID{0000-0003-4902-0486}}

\authorrunning{Y. Chen et al.}

%

\institute{Mechanical Engineering Department, University College London, London, UK \and Institute of Orthopaedic \& Musculoskeletal Science, University College London, London, UK \and Centre for Medical Image Computing, Wellcome/EPSRC Centre for Interventional and Surgical Sciences, Department of Medical Physics and Biomedical Engineering, University College London, London, UK \and
Royal National Orthopaedic Hospital, Stanmore, UK}
\maketitle              
\begin{abstract}
The recently proposed Segment Anything Model (SAM) is a general tool for image segmentation, but it requires additional adaptation and careful fine-tuning for medical image segmentation, especially for small, irregularly-shaped, and boundary-ambiguous anatomical structures such as the knee cartilage that is of interest in this work. Repaired cartilage, after certain surgical procedures, exhibits imaging patterns unseen to pre-training, posing further challenges for using models like SAM with or without general-purpose fine-tuning. To address this, we propose a novel registration-based prompt engineering framework for medical image segmentation using SAM. This approach utilises established image registration algorithms to align the new image (to-be-segmented) and a small number of reference images, without requiring segmentation labels. The spatial transformations generated by registration align either the new image or pre-defined point-based prompts, before using them as input to SAM. This strategy, requiring as few as five reference images with defined point prompts, effectively prompts SAM for inference on new images, without needing any segmentation labels. Evaluation of MR images from patients who received cartilage stem cell therapy yielded Dice scores of 0.89, 0.87, 0.53, and 0.52 for segmenting femur, tibia, femoral- and tibial cartilages, respectively. This outperforms atlas-based label fusion and is comparable to supervised nnUNet, an upper-bound fair baseline in this application, both of which require full segmentation labels for reference samples. The codes are available at: \url{https://github.com/chrissyinreallife/KneeSegmentWithSAM.git}
\end{abstract}
\section{Introduction}
\label{sec:intro}

Image segmentation forms the basis of various clinical tasks, including surgical planning \cite{alirr2018automated,litjens2014evaluation}, diagnostic assistance \cite{amyar2020multi,vorontsov2018liver}, and surgical navigation \cite{montana2021vessel}. Tissue inhomogeneity \cite{thippeswamy2021updates}, presence of various artefacts \cite{saeed2022image,saeed2022image_1} and operator interpretation variability \cite{czolbe2021segmentation,chalcroft2021development}, are challenges that often negatively impact segmentation performance of automated or manual segmentation methods. When coupled with limited dataset sizes in specialised clinical applications, due to labelling cost or low disease prevalence, it may become prohibitively infeasible to capitalise advances in modern machine learning~\cite{wang2019pathology}. 

\noindent\textbf{Atlas-based segmentation} first constructs one or more manually segmented images as a reference atlas library, before segmenting new patients by image registration between them. 
While being effective for brain imaging \cite{capobianco2023assessment}, the performance of atlas-based methods significantly diminishes for small and low-contrast soft tissue structures \cite{costea2022comparison}. 
The difficulty in hyaline cartilage segmentation, of interest in this work, arises from its thinness, relatively small size, curved morphology and challenging localisation of its interfaces (such as the interface between femoral and tibial cartilage) \cite{shan2014automatic}. 

\noindent\textbf{Learning-based methods}, such as U-Net \cite{ronneberger2015u}, V-Net \cite{milletari2016v} and thier variants, have achieved state-of-the-art performance for several medical image segmentation tasks \cite{siddique2021u}. However, 
their effectiveness is highly dependent on the labelled training data availability, often sparse and costly in the medical domain \cite{haque2020deep}, especially for pathological cases with their added complexity, variation and rarity~\cite{malhotra2022deep}. 
Few-shot segmentation methods can be adapted to new tasks with a few labelled samples, proposed for abdominal MR images \cite{tang2021recurrent}, skin images \cite{feyjie2020semi}, and pelvic MRs~\cite{li2023prototypical}. However, due to limited labelled data and significant differences in data distributions compared to scans with greater availability for training, application to many specialised areas, such as knee cartilage segmentation, remains a challenge \cite{ma2024segment,schreiner2021mocart}. Additionally, conventional few-shot learning approaches also necessitate large datasets, for cross-task learning \cite{wang2020generalizing}. 
To summarise, unique data distribution and extreme availability render existing learning-based approaches, arguably, inadequate for our segmentation problem \cite{shan2014automatic}.


\noindent\textbf{Vision foundation Models} such as SAM have emerged as general-purpose tools for image segmentation \cite{kirillov2023segment}. Building on these advances, the MedSAM foundation model, developed on a large-scale medical image dataset, allows for medical image segmentation for a variety of unseen clinical tasks \cite{ma2024segment}. 
While SAM and other similar models demonstrate superior generalisation compared to traditional few-shot segmentation approaches \cite{kirillov2023segment}, thanks to their greater training data diversity, they struggle to effectively segment rare out-of-training-data-distribution structures \cite{ma2024segment}. In addition, foundation models also enable segmentation through sparse annotation (such as selected points of interest) learnt from various training tasks, adaptable for out-of-distribution samples, with denser annotations leading to improved performance.
The sparse annotation at inference alleviates, to a degree, hurdles of data scarcity in medical imaging, however, it limits clinical applicability due to time and expertise constraints during inference.
To address this, prompt engineering approaches have been proposed 
\cite{huang2024learning}, but may still require large amounts of weakly-labelled or unlabeled images to learn effective prompting strategies. 
While these methods reduce annotation burden by replacing full manual segmentation with weak annotations, such as points and relatively more laborious bounding boxes, it remains an open question to develop effective prompting strategies for individual applications.

In this work, our proposed SAM prompting framework is based on spatially aligning the input image (or the point prompt) to a limited set of weakly-labelled reference samples, as opposed to training a prompt engineering network. 
This alignment mechanism enables obtaining segmentation for new samples without any manually annotated prompts for the sample itself. The weak annotations are only required for a small reference set of 5 images. 
As shown in the presented experimental results, the image/prompt alignment to a reference set enables accurate segmentation to be obtained from SAM, for out-of-distribution samples.
This method substantially reduces the data annotation cost associated with learning prompt engineering while enabling segmentation from foundation models using a limited set of weakly-labelled references, a practical scenario found in many potential clinical applications. 

The contributions of this work are summarised: 1) we introduce a SAM prompt engineering framework, based on aligning samples to-be-segmented with a small reference set of weakly-annotated samples, without the need for any large weakly-annotated dataset for learning prompting; 2) we propose two effective image registration strategies to align either images or prompts to the reference set of samples; 3) we demonstrate improved performance compared to alternative approaches which require full segmentation of the reference set and to using SAM without the proposed registration strategy, for a clinically challenging task of pathological bone tissue image segmentation.


\section{Methods}

\noindent Our proposed prompt engineering method for segmentation foundation models relies on an alignment of the input image or prompt to a limited reference set of weakly-labelled samples. This alignment allows us to construct effective prompts for segmentation foundation models such that we can obtain accurate segmentation, even for out-of-distribution samples, using as few as five point-prompt-labelled reference images.




\subsection{Segment anything model}

In this work, since we use a pre-trained, fixed-weight foundation model for segmentation, without any fine-tuning, the segmentation model is considered a non-parametric function:
$$f(x, p) = y: \mathcal{X} \times \mathcal{P} \rightarrow \mathcal{Y}$$
where $x\in\mathcal{X}$ is an image sampled from the image domain (considered out-of-distribution of the foundation model training image domain), $p\in\mathcal{P}$ is the prompt sampled from the domain of prompts, and $y\in\mathcal{Y}$ is a binary pixel-level segmentation label predicted by the non-parametric foundation model $f$. With a prompt $p_i$, a sample $x_i\in\mathcal{X}$ is segmented $y_i = f(x_i, p_i)$. In this work, the prompt is a set of $P$ points $p_i = \{p_{i, k}\}_{k=1}^P$ that specifies structures of interest in the segmentation (see examples in Fig.1).

\subsection{Reference and segmentation samples}

A set of $N$ reference images $\{x_i\}_{i=1}^N \in \mathcal{X}^N$ are used to form a reference set, with which to align the image that is to be segmented. For this reference set, we also have manually annotated prompts available $\{p_i\}_{i=1}^N \in \mathcal{P}^N$. 
For notational clarity, we define the image to be segmented, without any available prompts or labels, as $x_{new}\in \mathcal{X}$. The proposed approaches to obtain $x_{new}$ are outlined in the following sections.

\subsection{Image registration}

Image registration estimates a transform that warps a moving image to be aligned to a fixed image. The transform-predicting registration function is denoted as a non-parametric function:

$$
r(x_{\text{ref}}, x_{\text{new}}) = d: \mathcal{\mathcal{X}} \times \mathcal{X} \rightarrow \mathcal{D}
$$
where $x_{\text{ref}}\in\mathcal{X}$ is the reference image (used as the fixed image in this case), $x_{\text{new}}$ is the moving image (to be registered with the reference image), and $d\in\mathcal{D}$ is in spatial transformation domain $\mathcal{D}$ and warps the new image to be aligned with the reference image $x_{\text{ref}}$. the warped image $x'_{\text{new}}$ thus is:
$$
x'_{\text{new}} = x_{\text{new}} \circ d
$$

\subsection{Image alignment to the reference set}
\label{sec:method-image-align}

Aligning the new image $x_{new}$ to the reference $x_i$ is denoted as $x_{new} \circ d = x'_{new}$, where the transform is $d = r(x_i, x_{new})$. In this work, this transform is assumed invertable and the inverse of the transformation is thus denoted as $x'_{new} \circ (d)^{-1} = x_{new}$. For nonlinear transformation that is non-invertable, numerical approximation may be sought and further validated.

\begin{figure}
    \centering
     \includegraphics[width=1.0\textwidth]{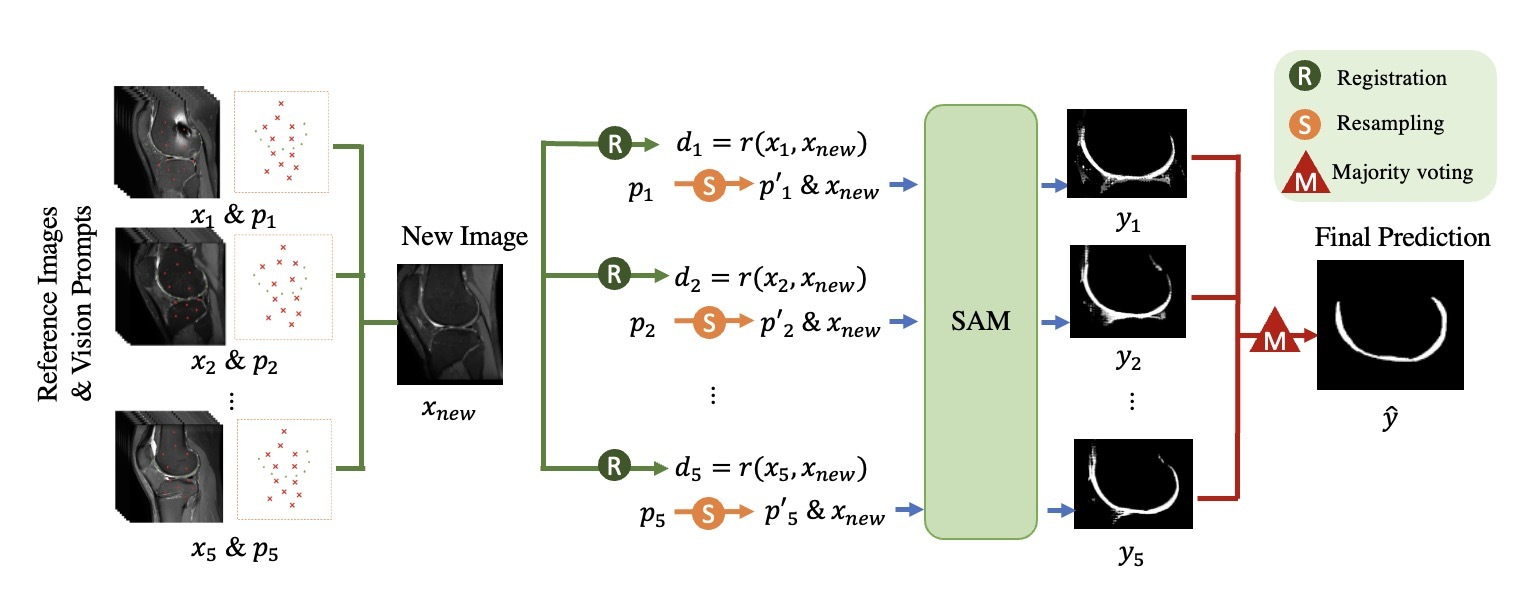}
    \caption{The proposed Image Alignment strategy, the first type of the two registration-enabled segmentation pipelines based on SAM prompt engineering, illustrated with samples of the reference images, point prompts and new images.}
    \label{fig1}
\end{figure}

\noindent\textbf{Obtaining pixel-level segmentation:} For the set of reference images $\{x_i\}_{i=1}^N$, we can compute the transforms to warp the image to be segmented $x_{new}$ to align with each of the reference images, denoted as $\{d_i\}_{i=1}^N = \{r(x_i, x_{new})\}_{i=1}^N$. We then obtain $N$ warped images $\{x'_{new, i}\}_{i=1}^N = \{x_{new} \circ d_i\}_{i=1}^N$.

The warped images $\{x'_{new, i}\}_{i=1}^N$ together with the original point prompts $\{p_i\}_{i=1}^N$, for the reference images, are used as input to the foundation model, in order to obtain a set of pixel-level segmentations $\{y_i\}_{i=1}^N = \{f(x'_{new, i}, p_i)\}_{i=1}^N$. We then apply the inverse transform to reverse the warp on these masks, denoted as $\{y'_i\}_{i=1}^N = \{y_i \circ (d_i)^{-1}\}_{i=1}^N$. A pixel-level majority voting \cite{nguyen2022topological} between these $N$ segmentations (after inverse transform applied), $\{y'_i\}_{i=1}^N$, then allows us to obtain the final prediction $\hat{y}$ for the sample $x_{new}$.

\begin{table}[!ht]
\centering
\renewcommand{\tablename}{Algorithm}
\caption{Segmentation Pipelines}\label{tab1}
\begin{tabular}{ll}
\hline
\bfseries (a) Image Alignment & \bfseries (b) Vision Prompt Alignment\\
\hline
{\bfseries Input:} New image $x_{new}$, &  {\bfseries Input:} New image $x_{new}$,\\
\indent\hspace{0.4cm}Reference images $\{x_i\}_{i=1}^N$, & \indent\hspace{0.4cm}Reference images $\{x_i\}_{i=1}^N$,\\
\indent\hspace{0.4cm}and Prompts $\{p_i\}_{i=1}^N$ & \indent\hspace{0.4cm}and Prompts $\{p_i\}_{i=1}^N$\\
1. Registration:  & 1. Registration: \\
\indent\hspace{0.4cm}$\{d_i\}_{i=1}^N= \{r(x_{new}, x_i)\}_{i=1}^N$ & \indent\hspace{0.4cm}$\{d_i\}_{i=1}^N= \{r(x_i, x_{new})\}_{i=1}^N$\\
2. Resampling: & 2. Resampling: \\
\indent\hspace{0.4cm}$\{x'_{new, i}\}_{i=1}^N = \{x_{new} \circ d_i\}_{i=1}^N$ & \indent\hspace{0.4cm}$\{p'_{i}\}_{i=1}^N = \{p_{i} \circ d_i\}_{i=1}^N$\\
3. SAM: $\{y_i\}_{i=1}^N= \{f(x'_{new}, p_i)_{i=1}^N$& 3. SAM: $\{y_i\}_{i=1}^N= \{f(x_{new}, p'_i)_{i=1}^N$\\

4. Resampling $\{y'_i\}_{i=1}^N = \{y_i \circ (d_i)^{-1}\}_{i=1}^N$ & 4. Majority Voting \\
5. Majority Voting& {\bfseries Output:} Final prediction $\hat{y}$ \\
{\bfseries Output:} Final prediction $\hat{y}$ & \\
\hline
\end{tabular}
\end{table}

\subsection{Prompt alignment to the new sample}
\label{sec:method-prompt-align}

In addition to the above-described mechanism of aligning the new image $x_{new}$ to a reference set $\{x_i\}_{i=1}^N$, for prompt construction, this section describes a second proposed mechanism of prompt construction, based on the same reference set, by aligning the prompts of the reference set $\{p_i\}_{i=1}^N$ to the new image $x_{new}$. 

To achieve prompt-alignment, image registration is again employed, similar to the image-alignment approach. However, instead of considering the reference samples as fixed images as in Sec.~\ref{sec:method-image-align}, we regard the new image to-be-segmented as the fixed image. Transforms are then obtained by warping the reference samples to the image to-be-segmented. The set of transforms that warp the reference samples $\{x_i\}_{i=1}^N$ to the image to be segmented $x_{new}$ is thus denoted as $\{d_i\}_{i=1}^N = \{r(x_{new}, x_i)\}_{i=1}^N$ (switched order of registration function inputs compared to image alignment).

\noindent\textbf{Obtaining pixel-level segmentation:} After computing the set of transforms $\{d_i\}_{i=1}^N = \{r(x_{new}, x_i)\}_{i=1}^N$ that warp the reference samples to the image to be segmented, we can, in this case, warp the point prompts $\{p_i\}_{i=1}^N$, from the reference set, using the computed transforms (instead of warping images as in the image-alignment strategy), which is denoted as $\{p'_{i}\}_{i=1}^N = \{p_i \circ d_i\}_{i=1}^N$. 

The image to be segmented $x_{new}$ along with warped point prompts $\{p'_{i}\}_{i=1}^N$ are then used as input to the foundation model, in order to obtain a set of pixel-level segmentations $\{y_i\}_{i=1}^N = \{f(x_{new}, p'_i)\}_{i=1}^N$. Similar to image-alignment, a pixel-level majority voting between these $N$ segmentations $\{y_i\}_{i=1}^N$ then allows us to obtain the final prediction $\hat{y}$ for the sample $x_{new}$.

\begin{figure}
    \centering
     \includegraphics[width=1.0\textwidth]{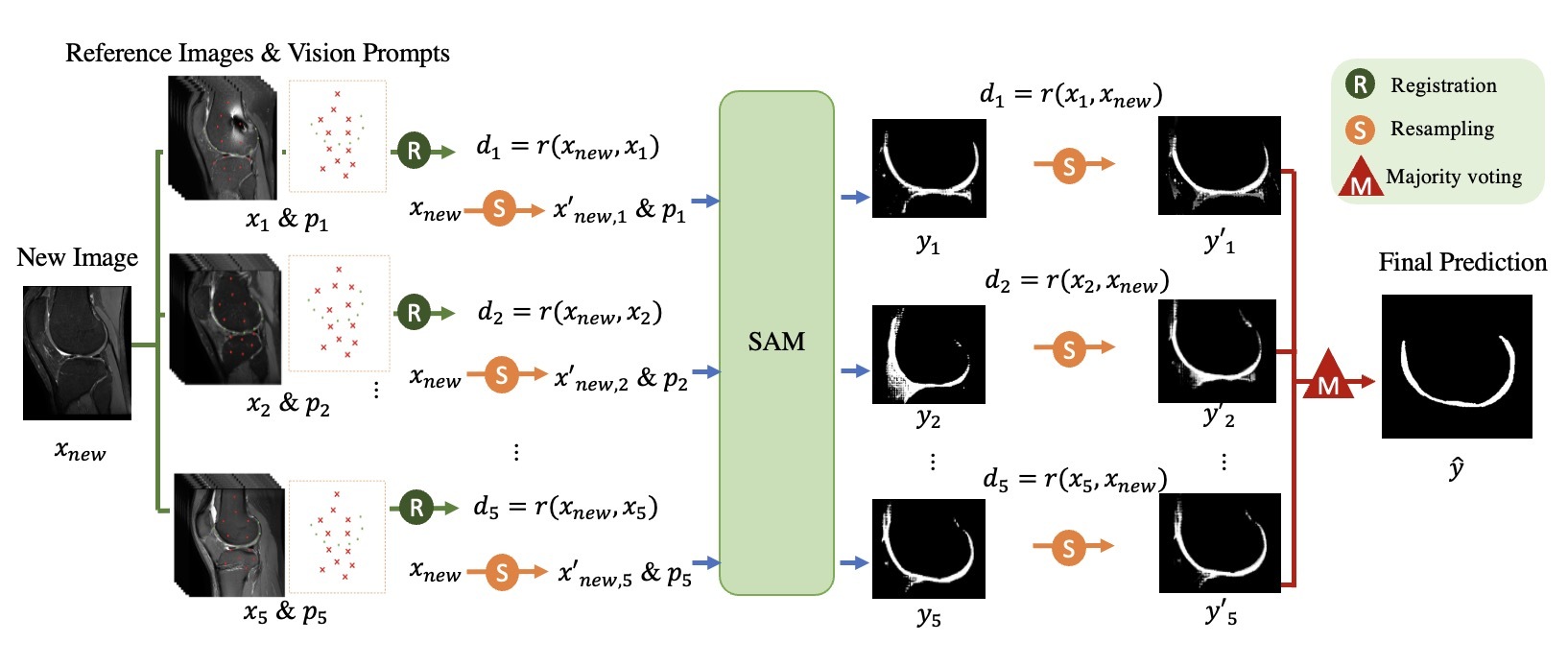}
    \caption{The proposed Prompt Alignment strategy, the second type of the two registration-enabled segmentation pipelines based on SAM prompt engineering, illustrated with samples of the new image, reference image and point prompts.}
    \label{fig2}
\end{figure}

\section{Experiments and Results}
\subsection{Dataset \& Implementation Details}
\textbf{Pathology Knee Joint MR data \& Pre-processing:} 
MR images were acquired from 35 patients who had undergone cartilage surgery with stem cell transplantation techniques at Royal National Orthopaedic Hospital~\cite{chimutengwende2021stem}.
Proton density (PD) weighted MR scans in the sagittal view were manually segmented to acquire anatomical and pathological labels, verified by an experienced radiology team. All images were normalized, resampled, and center-cropped to a consistent size of 512$\times$512$\times$40 and an image intensity range of [0, 3000].

\noindent\textbf{Vision prompts:} Out of the 35 samples, 5 were randomly chosen as reference images. Specific points prompts were manually selected for each reference image, guided by radiologists. These prompts encompass positive points (areas within the structure of interest) and negative points (areas outside the structure of interest). The same points prompt are used for comparisons between algorithms. Further details on these point prompts can be found in the open-source implementation of this project.


\noindent\textbf{Registration and Resampling:} 
In our study, NiftyReg \cite{modat2010fast} was utilized for both the registration and resampling processes. For registration, an initial affine registration, which aligns the images based on rotation, scale, and translation, was applied before the non-rigid algorithm, 
based on free-form deformation~\cite{rueckert1999nonrigid}.
The bending energy weight was set to 0.65, which empirically represent the stiffness characteristics of bone tissue in our experiments, with the sum-of-square difference (SSD) as the intra-modality similarity measure function. The remainder parameters were kept default values. These configurations, including the prior affine alignment and the choice of the well-established NiftyReg package, used in the available repository were based on preliminary experiments with qualitatively satisfactory alignment between the PD-weighted MR images.



\noindent\textbf{Post-registration point prompt filtering:}
A simple, automatic filtering was applied for point prompts, to ensure removal of mislabelled or misaligned prompt points, based on the intensity value of point prompt samples. For the femur and tibia, we used clipping intensity values [0, 1200] for positive points. For femoral and tibial cartilage, we used an intensity range of [800, 3000] for positive points. Any points lying outside these intensity ranges were removed. 
Pairs of images and corresponding prompts were then input into the default ViT-H SAM model to obtain the pixel-level segmentation, as described in Secs.~\ref{sec:method-image-align} and ~\ref{sec:method-prompt-align}.

\begin{table}[!ht]
\renewcommand{\thetable}{1}
    \centering
    \caption{Femur and Tibia segmentation results (top); Femoral Cartilage and Tibial Cartilage segmentation results (bottom). AR: usage of affine registration; NR: usage of non-rigid registration; FL: usage of full segmentation labels (opposed to point prompts, in our method). I-Align is our strategy based on image alignment and P-align is our strategy based on prompt alignment. Comparison needs to take into account the label requirement, as further discussed in the text.}
    \label{tab1}
    \begin{tabular}{|c  |c| c| c|  c| c| c| c|}
    \hline
    & & & & \multicolumn{2}{c}{\textbf{Femur}} & \multicolumn{2}{|c|}{\textbf{Tibia}} \\ 
    \cline{5-8} 
    \textbf{Method}         & AR & NR & FL              & \textbf{RMSD (mm)} & \textbf{Dice}  & \textbf{RMSD (mm)} & \textbf{Dice} \\ 
    \hline
    \textbf{Ours:I-align}   & \cmark & \cmark & \xmark  & $2.93\pm2.08$ & $0.89\pm0.04$ & $5.80\pm2.71$ & $0.87\pm0.03$ \\
    \textbf{Ours:P-align}   & \cmark & \cmark & \xmark  & $4.67\pm4.33$ & $0.86\pm0.09$ & $5.80\pm2.71$ & $0.87\pm0.03$ \\ 
    \hline
    I-align             & \cmark & \xmark & \xmark  & $5.64\pm3.82$ & $0.87\pm0.04$ & $5.37\pm2.72$ & $0.87\pm0.03$ \\
    P-align             & \cmark & \xmark & \xmark  & $5.02\pm4.34$ & $0.84\pm0.11$ & $4.11\pm4.90$ & $0.84\pm0.16$ \\ 
    NoReg                   & \xmark & \xmark & \xmark  & $6.37\pm4.98$ & $0.84\pm0.12$ & $3.19\pm1.39$ & $0.86\pm0.07$ \\
    \hline
    Atlas               & \cmark & \xmark & \xmark  & $7.71\pm8.59$ & $0.79\pm0.14$ & $6.18\pm6.70$ & $0.83\pm0.12$ \\
    Atlas               & \cmark & \cmark & \xmark  & $7.80\pm8.69$ & $0.79\pm0.14$ & $6.21\pm6.77$ & $0.83\pm0.12$ \\ 
    \hline
    nnUNet-2D               & - & - & \cmark  & $7.68\pm14.28$ & $0.82\pm0.29$ & $5.52\pm7.36$ & $0.83\pm0.20$ \\
    nnUNet-3D               & - & - & \cmark  & $15.50\pm22.45$ & $0.87\pm0.22$ & $29.56\pm33.22$ & $0.80\pm0.26$ \\ 
    \hline
    \multicolumn{8}{c}{}\\
    \hline
    & & & & \multicolumn{2}{c}{\textbf{Femoral Cartilage}} & \multicolumn{2}{|c|}{\textbf{Tibial Cartilage}}  \\ 
    \cline{5-8} 
    \textbf{Method}         & AR & NR & FL             & \textbf{VOE} & \textbf{Dice} & \textbf{VOE} & \textbf{Dice} \\ 
    \hline
    \textbf{Ours:I-align}   & \cmark & \cmark & \xmark  & $0.63\pm0.14$ & $0.53\pm0.15$ & $0.64\pm0.10$ & $0.52\pm0.11$ \\
    \textbf{Ours:P-align}   & \cmark & \cmark & \xmark  & $0.65\pm0.16$ & $0.50\pm0.18$ & $0.66\pm0.12$ & $0.50\pm0.14$ \\ 
    \hline
    I-align             & \cmark & \xmark & \xmark  & $0.66\pm0.14$ & $0.51\pm0.14$ & $0.67\pm0.07$ & $0.50\pm0.08$ \\
    P-align             & \cmark & \xmark & \xmark  & $0.65\pm0.16$ & $0.50\pm0.17$ & $0.68\pm0.11$ & $0.47\pm0.14$ \\ 
    NoReg                   & \xmark & \xmark & \xmark  & $ 0.60\pm0.03$ & $0.35\pm0.19$ & $0.92\pm0.08$ & $0.15\pm0.12$ \\
    \hline
    Atlas               & \cmark & \xmark & \xmark & $0.86\pm0.11$ & $0.22\pm0.17$ & $0.84\pm0.08$ & $0.26\pm0.12$ \\
    Atlas               & \cmark & \cmark & \xmark  & $0.76\pm0.54$ & $0.24\pm0.17$ & $0.84\pm0.07$ & $0.27\pm0.10$ \\ 
    \hline
    nnUNet-2D               & - & - & \cmark & $0.54\pm0.24$ & $0.59\pm0.27$ & $0.61\pm0.20$ & $0.52\pm0.24$ \\
    nnUNet-3D               & - & - & \cmark  & $0.46\pm0.21$ & $0.67\pm0.20$ & $0.49\pm0.13$ & $0.66\pm0.13$ \\ 
    \hline
    \end{tabular}
\end{table}

\subsection{Comparison and Ablation Experiments}
We compare our method with an atlas-based segmentation method \cite{rohlfing2005quo} and learning-based methods, 2D and 3D nnUNet~\cite{isensee2021nnu}. It is noteworthy that both the non-learning and learning-based approaches require full segmentation of the reference set of images and the same availability in labelled images reflects the real-world clinical scenarios as discussed in Sec.~\ref{sec:intro}. In this study, therefore, these are considered fair comparison between the proposed methods and the alternatives that have access to more detailed reference labels.

\noindent\textbf{Atlas-based segmentation:} In our implementation of the atlas-based method, all 5 reference images needed to be fully segmented. The 5 masks $\{y_{i}\}_{i=1}^N $ for the 5 reference images were warped from the reference space to the new image space (similar to the prompt-alignment strategy), denoted as $\{y'_{i}\}_{i=1}^{N=5} = \{y_i \circ d_i\}_{i=1}^N$. The final prediction $\hat{y}$ for $x_{new}$ was obtained by pixel-level majority voting \cite{nguyen2022topological}.

\noindent\textbf{Learning-based 2D\&3D nnUNet segmentation:}
For the purpose of providing a benchmark in this study, the dataset was divided into a training set comprising 200 images from 5 patients, for the comparison purpose based on the same label availability (but also requiring fully segmented labels), and the same testing set with 1200 images from 30 patients. The nnUNetv2 was employed \cite{isensee2021nnu}, leveraging both its 2D and 3D low resolution U-Net architectures.

\noindent\textbf{Ablation Studies:}
To assess the impact of various components of our proposed methods, we conducted ablation experiments, including configurations with or without affine registration (AR) and/or non-rigid registration (NR). The benefits of increased registration accuracy appear to diminish, in terms of achieving improved segmentation performance. Varying the number of reference images was also tested, with fewer reference images leading to a poorer segmentation performance as expected. Average dice scores of 0.11 for cartilage and 0.24 for the femur were obtained using only 3 reference images. More reference images lessen the label-to-performance efficiency, as can be verified using our open-sourced implementation and provided example images. However, future study may be interesting to explore the relationships between the required or optimal reference set size and the clinically required accuracy.


\subsection{Results based on Clinically-Reported Evaluation Metrics}
A root-mean-square symmetric surface distance (RMSD) \cite{heimann2010segmentation} is reported for bone segmentation. An additional volume overlap error (VOE), commonly reported for cartilage segmentation, is also included. All metrics were computed over slices containing bone tissue. The Dice coefficient is also reported for all experiments. These metrics were adopted for our post-procedure evaluation in current clinical and research practice for cartilage repair, where criteria such as MOCART (Magnetic Resonance Observation of Cartilage Repair Tissue) are based on the measurements on bone tissue surfaces, volumes and other morphological features.

Our method, with only five weakly-labelled reference images, achieves Dice scores of 0.89 and 0.87, for the femur and tibia segmentation, respectively. Comparatively lower Dice scores of 0.82 and 0.83, respectively, were observed for the nnUNet-2D.
For femoral and tibial cartilage, the Dice scores of our proposed methods are 0.53 and 0.52, closely matching or slightly underperforming against nnUNet-2D's 0.59 and 0.52, which requires full segmentation labels for the reference images. In cases, such as tibia segmentation, our method yielded a superior Dice coefficient but relatively average RMSD, as shown in Table~\ref{tab1}. This discrepancy may be due to the higher anatomical variability of the tibia between patients, whereas the proposed approach may prioritise overlap over edge detection, for achieving a globally less biased segmentation, which may worth quantifying for further understanding. The ANOVA test followed by Tukey HSD among all the methods confirm our method significantly surpasses atlas-based methods in bone and cartilage segmentation with p-values of 0.034 and 0.023 respectively, as well as ablated variants for cartilage segmentation. Further results are summarised in Table~\ref{tab1}.

\section{Discussion and Conclusion}

We show statistically significant performance improvements for our method, compared to other tested methods and to ablated variants, demonstrating the utility of our framework for segmenting new images using only five weakly-labelled references. In contrast, other methods with comparable performance utilise full segmentation labels - a non-trivial expertise-demanding task even for a small reference set due to the relatively irregularly-shaped structures. It is worth noting that lower performance was observed in cartilage segmentation in our method compared to the nnUNet approach which uses full segmentation labels, which may require further test data set to show statistical significance and, if significant, can potentially be bridged by further development of the proposed approaches. Cartilage areas are typically curved, elongated and small, which makes them inherently difficult to segment accurately using only weak labels. This complexity suggests that additional information provided by full segmentation annotations could be beneficial. Future research could explore incorporating such supplementary data to address these challenges. Additionally, a targeted registration concentrating on bone and cartilage interfaces instead of the whole image, could potentially enhance performance.

Our results, for a real clinical patient dataset for knee MR segmentation, demonstrate higher performance and, perhaps more importantly, a more practical and feasible segmentation approach, compared to existing methods. The improved performance demonstrates the effectiveness of utilising weakly-labelled point prompts in combination of well-established image registration algorithms within our proposed framework, with potentials especially in many real-world clinical applications in which datasets can be extremely constrained by the availability of pathological images and pixel-level annotated samples.

\section{Acknowledgement}
This work was supported in part by EPSRC [EP/T029404/1], Wellcome/EPSRC Centre for Interventional and Surgical Sciences [203145Z/16/Z] and the International Alliance for Cancer Early Detection, a partnership between Cancer Research UK [C73666/A31378], Canary Center at Stanford University, the University of Cambridge, OHSU Knight Cancer Institute, University College London and the University of Manchester.



\newpage
%
%
%
\bibliographystyle{splncs04}
\bibliography{reference}
%





\end{document}